\def\year{2017}\relax
\documentclass[letterpaper]{article}
\usepackage{aaai17}
\usepackage{times}
\usepackage{helvet}
\usepackage{url}
\usepackage{courier}
\usepackage{amsfonts}
\usepackage{amsmath}
\usepackage{color}
\usepackage{graphicx}
\usepackage{tabularx}
\usepackage{multirow}
\usepackage{paralist}
\usepackage{graphicx}
\begin{document}
%
\title{Joint Copying and Restricted Generation for Paraphrase
	}
\author{
	Ziqiang Cao$^{1,2}$  ~~ ~~ Chuwei Luo$^3$ ~~ ~~ Wenjie Li$^{1,2}$ ~~ ~~ Sujian Li$^4$\\
	$^1$Department of Computing, The Hong Kong Polytechnic University, Hong Kong\\
	$^2$Hong Kong Polytechnic University Shenzhen Research Institute, China \\
	$^3$School of Computer Science, Wuhan University, China\\
	$^4$Key Laboratory of Computational Linguistics, Peking University, MOE, China \\
	{\tt \{cszqcao, cswjli\}@comp.polyu.edu.hk  } \\
	{\tt luochuwei@whu.edu.cn} \\
	{\tt lisujian@pku.edu.cn} \\
}
\maketitle
\begin{abstract}
\begin{quote}

Many natural language generation tasks, such as abstractive summarization and text simplification, are paraphrase-orientated.
In these tasks, copying and rewriting are two main writing modes. 	
Most previous sequence-to-sequence (Seq2Seq) models use a single decoder and neglect this fact.
In this paper, we develop a novel Seq2Seq model to fuse a copying decoder and a restricted generative decoder.
The copying decoder finds the position to be copied based on a typical attention model.
The generative decoder produces words limited in the source-specific vocabulary.
To combine the two decoders and determine the final output, we develop a predictor to predict the mode of copying or rewriting.
This predictor can be guided by the actual writing mode in the training data.
We conduct extensive experiments on two different paraphrase datasets.
The result shows that our model outperforms the state-of-the-art approaches in terms of both informativeness and language quality.	
\end{quote}
\end{abstract}

\section{Introduction}
Paraphrase is a restatement of the meaning of a text using other words.
Many natural language generation tasks are paraphrase-orientated.
For example, abstractive summarization is to use a condensed description to summarize the main idea of a document, while text simplification is to simplify the grammar and vocabulary of a document.
In paraphrase, copying and rewriting are two main writing modes. 
Recently, the encoder-decoder structure (aka. SeqsSeq model) has become more and more popular in many language generation tasks \cite{bahdanau2014neural,shang2015neural}.
In such a structure, the source text is encoded by the encoder as a context vector.
Then, a decoder decodes the semantic information in the vector and outputs the target text.
Studies such as \cite{rush-chopra-weston:2015:EMNLP,hu2015lcsts} have applied the popular SeqsSeq model initially used in machine translation~\cite{bahdanau2014neural} to the paraphrase task.
Despite the competitive performance, these models seldom take into account the two major writing modes of paraphrase.
 
On the one hand, due to the nature of the task, keywords of the source text are usually reserved in the target text.
However, with only one decoder generating over the entire vocabulary, a typical Seq2Seq model fails to reflect the copying mode.
As a result, many keywords provided in the source text may be overlooked in the target text.
In addition, certain keywords like named entities are often rare words and  masked as unknown (UNK) tags in Seq2Seq models, which unavoidably causes the decoder to generate a number of UNK tags. 
Although not aiming to explicitly explore the copying mechanism, the work of \cite{rush-chopra-weston:2015:EMNLP} finds that it largely improves the performance to add the input-related hand-crafted features to guide the generation.


On the other hand, rewriting also plays a significant role in paraphrase.
In this writing mode, although the target words are not the same as the source words, there are semantic associations between them.
For example, ``seabird'' is possibly generalized as ``wildlife'' in summarization, and ``the visually impaired'' can be converted into ``people who can not see'' in text simplification.
The decoders in most previous work generate words by simply picking the likely target words that fit the contexts out of a large vocabulary.
This common practice suffers from two problems.
First, the computation complexity is linear to the vocabulary size.
In order to cover enough target words, the vocabulary size usually reaches $10^4$ or even $10^5$.
Consequently, the decoding process becomes quite time-consuming.
Moreover, the decoder sometimes produces the named entities or numbers which are common but do not exist in or even irrelevant to the input text, and in turn the meaningless paraphrases.

In this paper, we develop a novel Seq2Seq model called CoRe, which captures the two \textbf{core} writing modes in paraphrase, i.e., \textbf{Co}pying and \textbf{Re}writing.
CoRe fuses a copying decoder and a restricted generative decoder.
Inspired by \cite{vinyals2015pointer}, the copying decoder finds the position to be copied based on the existing attention mechanism.
Therefore, the weights learned by the attention mechanism have the explicit meanings in the copying mode.
Meanwhile, the generative decoder produces the words restricted in the source-specific vocabulary.
This vocabulary is composed of a source-target word alignment table and a small set of frequent words.
The alignment table is trained in advance, and many frequent rewriting patterns are included in it.
It seems better to update the alignment table according to the learned attention weights.
However, with the supplement of a few frequent words, experiments (see Table:\ref{tb:coverage}) show that more than 95\% of the target words have already been covered by our decoders.
While the output dimension of our generative decoder is just one tenth of the output dimension used by the common Seq2Seq models, it is able to generate highly relevant words concerning rewriting.
To combine the two decoders and determine the final output, we develop a predictor to predict the writing mode of copying or rewriting.
Since we know the actual mode at each output position in a training instance, we introduce a binary sequence labeling task to guide the learning of this predictor, which takes advantages of the supervision derived from the writing modes.

To the best of our knowledge, the work most relevant to ours is the COPYNET~\cite{gu2016incorporating} which also explores the copying mechanism.
However, COPYNET adds an additional attention-like layer to predict the copying weight distribution.
This layer then competes with the output of the generative decoder.
Therefore, it is not easy for COPYNET to explain the contributions of copying and generation.
Compared with our model, COPYNET introduces a lot of extra parameters and ignores the supervision derived from the writing modes.
Moreover, the generative decoder of COPYNET is only allowed to produce frequent words.
As a result, the rewriting patterns are discarded to a large extent.

We conduct extensive experiments on two different paraphrase tasks, i.e., abstractive summarization and text simplification.
The result shows that both informativeness and sentence quality of our model outperform the state-of-the-art Seq2Seq models as well as the statistical machine translation approaches.

The contributions of our work are as follows:
\begin{itemize}
	\item We develop two different decoders to simulate major human writing behaviors in paraphrase.
	\item We introduce a binary sequence labeling task to predict the current writing mode, which utilizes additional supervision.
	\item We add restrictions to the generative decoder in order to produce highly relevant content efficiently.
\end{itemize}


\section{Background: Seq2Seq Models and Attention Mechanism}
Seq2Seq models have been successfully applied to a series of natural language generation tasks, such as machine translation~\cite{bahdanau2014neural}, response generation~\cite{shang2015neural} and abstractive summarization~\cite{rush-chopra-weston:2015:EMNLP}.
With these models, the source sequence $X = [{x_1}, \cdots ,{x_n}]$ is converted into a fixed length context vector ${\mathbf{c}}$, usually by a Recurrent Neural Network (RNN) encoder, i.e.,
\begin{align}
{{\mathbf{h}}_\tau} &= f({x_\tau},{{\mathbf{h}}_{\tau - 1}}) \\
{\mathbf{c}} &= \phi ({{\mathbf{h}}_1}, \cdots {{\mathbf{h}}_n}) \label{eq:ct}
\end{align}
where $\{ {{\mathbf{h}}_\tau}\}$ are the RNN states, $f$ is the dynamics function, and $\phi$ summarizes the hidden states, e.g., choosing the last state $\mathbf{h}_n$. 

The decoder unfolds the context vector ${\mathbf{c}}$ into the target RNN state ${\mathbf{s}}_t$ through the similar dynamics in the encoder:
\begin{equation} \label{eq:st}
{{\mathbf{s}}_t} = f({y_{t - 1}},{{\mathbf{s}}_{t - 1}},{\mathbf{c}}) 
\end{equation}
Then, the predictor is followed to generate the final sequence, usually using a softmax classifier:
\begin{equation}\label{eq:predict}
p({y_t}|{y_{ < t}},{\mathbf{X}}) = \frac{{\exp ({\mathbf{w}}_t\psi ({y_{t - 1}},{{\mathbf{s}}_t},{{\mathbf{c}}_t}))}}{{\sum\nolimits_{{y_{t'}} \in {\mathbf{V}}} {\exp ({\mathbf{w}}_{t'}\psi ({y_{t - 1}},{{\mathbf{s}}_t},{{\mathbf{c}}_t}))} }}
\end{equation}
where $y_t$ is the predicted target word at the state $t$, ${\mathbf{w}}_t$ is the corresponding weight vector, and $\psi$ is an affine transformation.
$\mathbf{V}$ is the target vocabulary, and it is usually as large as $10^4$ or even $10^5$.

To release the burden of summarizing the entire source into a single context vector, the attention mechanism~\cite{bahdanau2014neural} uses a dynamically changing context $\mathbf{c}_t$ to replace $\mathbf{c}$ in Eq.~\ref{eq:st}.
A common practice is to represent $\mathbf{c}_t$ as the weighted sum of the source hidden states:
\begin{align}
{\alpha _{t\tau }} &= \frac{{{e^{\eta ({{\mathbf{s}}_{t - 1}},{{\mathbf{h}}_\tau })}}}}{{\sum\nolimits_{\tau ' = 1}^n {{e^{\eta ({{\mathbf{s}}_{t - 1}},{{\mathbf{h}}_{\tau '}})}}} }}\quad ,\forall \tau  \in [1,n]  \label{eq:align} \\
{{\mathbf{c}}_t} &= \sum\nolimits_{\tau  = 1}^n {{\alpha _{t\tau }}{{\mathbf{h}}_\tau }} 
\end{align}
where $\alpha _{t\tau }$ reflects the alignments between source and target words, and $\eta$ is the function that shows the correspondence strength for attention, usually approximated with a deep neural network.

\section{Method}

\begin{figure*}
\centering
\includegraphics[width=0.7\linewidth]{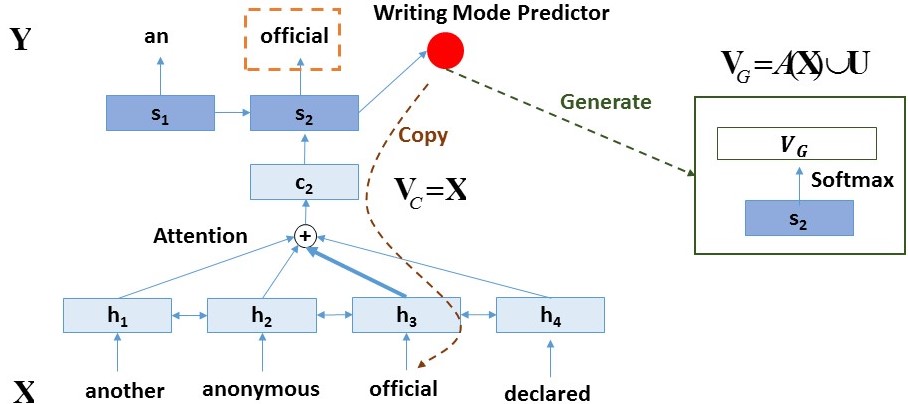}
\caption{Model overview.}
\label{fig:model}
\end{figure*}

As illustrated in Figure~\ref{fig:model}, CoRe is based on the encoder-decoder structure.
The source sequence is transformed by a RNN \textbf{Encoder} into the context representation, which is then read by another RNN \textbf{Decoder} to generate the target sequence.

\subsection{Encoder}
We follow the work of \cite{bahdanau2014neural} to build the encoder.
Specifically, we use the Gated Recurrent Unit (GRU) as the recurrent unit, which often performs much better than the vanilla RNN.
The bi-directional RNN is introduced to make the hidden state $\mathbf{h}_\tau$ aware of the contextual information from both ends.
Then, we use the attention mechanism to build the context vector as Eq.~\ref{eq:ct}.
However, unlike most previous work, we re-use the learned alignments (Eq.~\ref{eq:align}) in the decoders.

\subsection{Decoder}
Instead of using the canonical RNN-decoder like \cite{bahdanau2014neural}, we develop two distinct decoders to simulate the copying and rewriting behaviors, respectively.

The \textbf{Copying Decoder} ($C$) picks the words from the source text.
In paraphrase, most keywords from the original document will be reserved in the output.
This decoder captures this fact. 
Since the attention mechanism is supposed to provide the focus of source text during generation, for the copying behavior, the weights learned by Eq.~\ref{eq:align} can be interpreted as the copying probability distribution.
Therefore, the output of the copying decoder is as follows:
\begin{equation}\label{eq:p_copy}
{p_C}({y_t}|{y_{ < t}},{\mathbf{X}}) = \left\{ {\begin{array}{*{20}{l}}
	{{\alpha _{t\tau }},\quad \text{if }{y_t} = {x_\tau }} \\ 
	{0,\quad   \text{otherwise}} 
	\end{array}} \right.
\end{equation}
Most previous work uses the attention mechanism as a module to build the context vector.
In contrast, our copying decoder provides the explicit meanings (i.e., the copying probability distribution) to the learned alignment.
Notice that, this decoder only generates words in the source, i.e., the vocabulary for this decoder is $\mathbf{V}_C=\mathbf{X}$.
We observe that quite a number of low-frequency words in the actual target text are extracted from the source text.
Therefore, the copying decoder largely reduces the chance to produce the unknown (UNK) tags.

The \textbf{Restricted Generative Decoder} ($G$), on the other hand, restricts the output in a small yet highly relevant vocabulary according to the source text.
Specifically, we train a rough alignment table $\mathbf{A}$ based on the IBM Model~\cite{dyer2013simple} beforehand.
This table is able to capture many representative rewriting patterns, such as ``sustain $\to$ injury'', and ``seabird $\to$ wildlife''.
Our pilot experimental results show that the alignment table covers most target words which are not extracted from the source.
To further increase the coverage, we supplement an additional frequent word table ${\mathbf{U}}$\footnote{We add the UNK tag in this table}.
Putting together, the final vocabulary for this decoder is limited to:
\begin{equation}\label{eq:v_g}
{{\mathbf{V}}_G} = {\mathbf{A}}({\mathbf{X}}) \cup {\mathbf{U}}
\end{equation}
In our experiments, we retain 10 most reliable alignments for a source word, and set $|\mathbf{U}|=2000$.
As a result, ${{\mathbf{V}}_G}$ is only one tenth of the vocabulary $\mathbf{V}$ used by the common canonical RNN-decoder in size.
The output of this decoder is formulated by Eq.~\ref{eq:predict2} which is similar to Eq.~\ref{eq:predict}, except for the reduced vocabulary:
\begin{equation}\label{eq:predict2}
p_G({y_t}|{y_{ < t}},{\mathbf{X}}) = \frac{{\exp ({\mathbf{w}}_t^T\psi ({y_{t - 1}},{{\mathbf{s}}_t},{{\mathbf{c}}_t}))}}{{\sum\nolimits_{{y_{t'}} \in {\mathbf{V}_G}} {\exp ({\mathbf{w}}_{t'}^T\psi ({y_{t - 1}},{{\mathbf{s}}_t},{{\mathbf{c}}_t}))} }}
\end{equation}
Compared with the generation on the large vocabulary $\mathbf{V}$, the restricted decoder not only runs much faster but also produces more relevant words.

To combine the two decoders, we introduce a binary sequence labeling task to decide whether the current target word should come from copying or rewriting.
Specifically, for each hidden state $\mathbf{s}_t$, we compute a predictor $\lambda _t$ to represent the probability of copying at the current generation position:
\begin{equation}\label{eq:lambda}
{\lambda _t} = \sigma ({{\mathbf{w}}_C}{{\mathbf{s}}_t})
\end{equation}
where $\sigma$ is the sigmoid function and ${\mathbf{w}}_C$ is the weight parameters.
$\lambda _t$ measures the contributions of the two decoders, and the final combined prediction probability is:
\begin{equation} \label{eq:output}
p({y_t}|{y_{ < t}},{\mathbf{X}}) = {\lambda _t}{p_C} + (1 - {\lambda _t}){p_G}
\end{equation}
It is noted that $\lambda _t$ has the following actual supervision in the training set:
\begin{equation}\label{eq:lambda_supervision}
\lambda _t^* = \left\{ {\begin{array}{*{20}{l}}
	1,\quad \text{if target word at }t \text{ exists in the source} \\ 
	0,\quad \text{otherwise} 
	\end{array}} \right.
\end{equation}
Therefore, we can utilize this supervision to guide the writing mode prediction.

The common canonical RNN-decoder outputs the probability distribution over the reserved target word vocabulary $\mathbf{V}$.
Since the computation complex of a Seq2Seq model is linear to the output dimension of the decoder, a large amount of infrequent target words have to be discarded to ensure a reasonable vocabulary size.
As a result, a target sentence may contain many UNK tags, and thus unreadable.
By contrast, the output dimension of our generative decoder is totally independent on the reserved target word vocabulary.
Therefore, we opt to reserve all the target words in the training set.
Experiments demonstrate that our model runs efficiently and rarely generates the UNK tags.

\subsection{Learning}
The cost function $\epsilon$ in our model is the sum of two parts, i.e.,
\begin{equation}
\epsilon=\epsilon _1+\epsilon _2
\end{equation}
The first one $\epsilon _1$ is the difference between the output $\{y_t\}$ and the actual target sequence $\{y_t^*\}$.
As the common practice, we use Cross Entropy (CE) to measure the difference of probability distributions:
\begin{equation}\label{eq:cost1}
{\varepsilon _1} = - \sum\nolimits_t {\ln (p(y_t^*|{y_{ < t}},{\mathbf{X}}))} 
\end{equation}
In most existing Seq2Seq models, $\epsilon _1$ is the final cost function.
However, in our model, we include another cost function $\epsilon _2$ derived from the prediction of writing modes.
As shown in Eq.~\ref{eq:lambda}, a binary sequence labeling process in our model predicts whether or not the current target word is copied.
$\epsilon _2$ measures the performance in this task,
\begin{equation}\label{eq:cost2}
{\varepsilon _2} =-( \sum\nolimits_t {(\lambda _t^*\ln ({\lambda _t})}  + (1 - \lambda _t^*)\ln (1 - {\lambda _t})))
\end{equation}
$\epsilon _2$ utilizes the additional supervision of the training data.
The experiments show that this cost function accurately balances the proportion of the words derived from copying and generation.

Given the cost function $\epsilon$, we use the RmsProp~\cite{tieleman2012lecture} optimizer with mini-batches to tune the model weights.
RmsProp is a popular method to train recurrent neural networks.

\section{Experiments}

\subsection{Dataset}
We test our model on the following two paraphrase-orientated tasks,
\begin{enumerate}
	\item One-sentence abstractive summarization
	\item Text simplification
\end{enumerate}
One-sentence summarization is to use a condensed sentence (aka. highlight) to describe the main idea of a document.
This task facilitates efficient reading. 
Text simplification modifies a document in such a way that the grammar and vocabulary is greatly simplified, while the underlying meaning remains the same.
It is able to make the scientific documents easily understandable for outsiders.
We build datasets for both tasks based on the existing work.

\textbf{One-sentence Abstractive Summarization:}
For this task, we need a corpus that consists of $<$document,highlight (one-sentence summary)$>$ pairs.
We modify an existing corpus that has been used for the task of passage-based question answering~\cite{hermann2015teaching}.
In this work, a collection of news documents and the corresponding highlights are downloaded from CNN and Daily Mail websites. 
For each highlight, we reserve the original sentences that have at least one word overlap with the source text.
Therefore, if a document holds multiple highlights, the source text for each highlight can be different.

  
\textbf{Text Simplification:}
Simple English Wikipedia\footnote{\url{http://simple.wikipedia.org}} articles represent a simplified version of traditional English Wikipedia articles.
\cite{kauchak2013improving} built a $<$Wikipedia text, Simple English Wikipedia text$>$ corpus according to the aligned articles.
We eliminate the non-English words in the corpus, and remove the pairs where the source and the target are exactly the same.

The basic information of the two datasets are presented in Table~\ref{tb:datasize}.
As can be seen, each dataset has a large vocabulary size.
In the summarization dataset, the target length is much shorter than the source length, while in the simplification dataset, their lengths are similar.

In addition, we compute the target word coverage ratio based on different vocabulary sets, as shown in Table ~\ref{tb:coverage}.
It appears that both datasets hold a high copying ratio.
When we restrict the generative decoder to produce the source alignments, more than 85\% target words can be covered.
When combined with 2000 frequent words, the coverage ratio of our model is already close to that using the vocabulary of 30000 words. 

\begin{table}[ht]
	\centering
	\small
	\begin{tabular}{l|ll}
		\hline
		Statistics       & Summarization & Simplification \\ \hline
		Training\#    & 986637        & 132609         \\
		Validation\#  & 51759         & 6700           \\
		Test\#        & 42003         & 3393           \\
		Source Length & 71.0          & 24.3           \\
		Target Length & 12.6          & 20.9           \\
		Vocab Size    & 116130        & 123304         \\ \hline
		\end{tabular}
		\caption{Statistics of the two datasets.}
		\label{tb:datasize}
		\end{table}

\begin{table}[ht]
	\centering
	\small
	\begin{tabular}{l|ll}
		\hline
		Vocabulary    & Summarization & Simplification \\ \hline
		$\mathbf{X}$     & 79.2          & 78.1           \\
		$\mathbf{X}\cup \mathbf{A}(\mathbf{X})$         & 89.2          & 85.8           \\
		$\mathbf{X}\cup\mathbf{A}(\mathbf{X})\cup\mathbf{U}$      & 95.3          & 96.0           \\ 
		$|\mathbf{V}|=30000$ & 96.3          & 95.4           \\ \hline
	\end{tabular}
	\caption{Target word coverage ratio (\%) on the test set. }
	\label{tb:coverage}
\end{table}

\subsection{Implementation}
Turned on the validation dataset, we set the dimension of word embeddings to 256, and the dimension of hidden states to 512.
The initial learning rate is 0.05 and the batch size is 32.
Our implementation is based on the standard Seq2Seq model dl4mt\footnote{\url{https://github.com/nyu-dl/dl4mt-multi}} under the Theano framework\footnote{\url{http://deeplearning.net/software/theano/}}.
We leverage the popular tool Fast Align~\cite{dyer2013simple} to construct the source-target word alignment table $\mathbf{A}$.
The vocabulary of our generative decoder is restricted in the top 10 alignments of the source words plus 2000 frequent words.
Although our model is more complex than the standard attentive Seq2Seq model, it only spends two thirds of the time in both training and test.

\subsection{Evaluation Method}
Informativeness is evaluated using ROUGE\footnote{ROUGE-1.5.5 with options: -n 2 -m -u -c 95 -x -r 1000 -f A -p 0.5 -t 0.}~\cite{lin2004rouge}, which has been regarded as a standard automatic summarization evaluation metric.
ROUGE counts the overlapping units such as the n-grams, word sequences and word pairs between the candidate text $\mathbf{Y}$ and the actual target text $\mathbf{T}$. 
As the common practice, we take ROUGE-1 and ROUGE-2 scores as the main metrics.
They measure the uni-gram and bi-gram similarities, respectively.
For example, the f-score of ROUGE-2 is computed as follows:
\begin{align}
\text{ROUGE} - {2_{{\text{f - score}}}} = \frac{{2 \times \sum\nolimits_{b \in {\mathbf{Y}}} {\min \{ {N_{\mathbf{Y}}}(b),{N_{\mathbf{T}}}(b)\} } }}{{\sum\nolimits_{b \in {\mathbf{Y}}} {{N_{\mathbf{Y}}}(b)}  + \sum\nolimits_{b \in {\mathbf{T}}} {{N_{\mathbf{T}}}(b)} }}
\end{align}
where $b$ stands for a bi-gram.
${N_{\mathbf{Y}}}(b)$, ${N_{\mathbf{T}}}(b)$ are the numbers of the bi-gram $b$ in the candidate text and target text, respectively.
Since we do not try to control the length of the generated sentences, we use the f-score rather than the recall for comparison.

We conduct text quality evaluation from several points of view.
We use SRILM\footnote{\url{http://www.speech.sri.com/projects/srilm/}} to train a 3-gram language model on the entire target datasets, and compute the perplexity (PPL) of the generated text.
The lower PPL usually means higher readability.
We also perform the statistical analysis on the average length of the target text, UNK ratio and copy ratio.
We assume that the good sentences ought to have the similar length and copying ratio to the answers (refer to Table~\ref{tb:datasize} and \ref{tb:coverage}), and their UNK ratio should be low.

\subsection{Baselines}
We compare the proposed model CoRe with various typical methods.
At first, we introduce the standard baseline called ``LEAD''.
It simply selects the ``leading'' words from the source as the output.
According to the averaged target length in Table~\ref{tb:datasize}, we 
choose the first 20 words for summarization and 25 for simplification.
We also introduce the state-of-the-art statistical machine translation system Moses~\cite{koehn2007moses} and the Seq2Seq model ABS~\cite{rush-chopra-weston:2015:EMNLP}.
Moses is the dominant statistical approach to machine translation. 
It takes in the parallel data and uses co-occurrence of words and phrases to infer translation correspondences.
For fair comparison, when implementing Moses, we also employ the alignment tool Fast Align and the language model tool SRILM.
ABS is a Seq2Seq model with the attention mechanism.
It is similar to the neural machine translation model proposed in \cite{bahdanau2014neural}.
ABS has achieved promising performance on another one-sentence summarization benchmark.

Note that, we would like to but fail to take COPYNET~\cite{gu2016incorporating} into comparison.
Its source code is not publicly available. 

\subsection{Performance}
\begin{table*}[ht]
	\centering
	\small
	\begin{tabular}{l|l|ll|llll}
		\hline
		\multirow{2}{*}{Data}           & \multirow{2}{*}{Model} & \multicolumn{2}{c|}{Informativeness} & \multicolumn{4}{c|}{Text Quality}            \\ \cline{3-8} 
		&                        & ROUGE-1(\%)       & ROUGE-2(\%)      & PPL           & Length & UNK(\%) & Copy(\%) \\ \hline
		\multirow{4}{*}{Summarization}  & LEAD                   & 28.1              & 14.1             & 176           & 19.9   & 0       & 100      \\
		& Moses                  & 27.8              & 14.1             & 214           & 73.0   & 0$^*$       & 99.6     \\
		& ABS                    & 28.1              & 12.4             & 113           & 13.7   & 0.88    & 92.0     \\
		& CoRe                     & \textbf{30.5}     & \textbf{16.2}    & \textbf{95}   & 14.0   & 0.14    & 88.6     \\ \hline
		\multirow{4}{*}{Simplification} & LEAD                   & 66.4              & 49.4             & 66.5          & 20.8   & 0       & 100      \\
		& Moses                  & 70.9              & 52.1             & 70.3          & 24.4   & 0$^*$       & 97.6     \\
		& ABS                    & 68.4              & 50.3             & 69.5          & 22.7   & 5.6     & 87.7     \\
		& CoRe                     & \textbf{72.7}     & \textbf{55.3}    & \textbf{60.9} & 19.6   & 2.3     & 85.9     \\ \hline
	\end{tabular}
	\caption{Performance of different models. $^*$Moses simply ignore the unknown words.}
	\label{tb:result}
\end{table*}

The results of different approaches are presented in Table~\ref{tb:result}.
In this table, the metrics that measure informativeness and text quality are separated.
Let's look at the informativeness performance first.
As can be seen, CoRe achieves the highest ROUGE scores on both summarization and text simplification.
In contrast, the standard attentive Seq2Seq model ABS is slightly inferior to Moses.
It even performs worse than the simple baseline LEAD in terms of ROUGE-2 in summarization.
Apparently, introducing the copying and restricted generation mechanisms is critical for the paraphrase-oriented tasks.

Then, we check the quality of the generated sentences.
According to PPL, the sentences produced by CoRe resemble the target language the most.
It is interesting that LEAD extracts human-written text in the source.
Nevertheless, its PPL is considerably higher than CoRe on both datasets.
It seems that CoRe indeed captures some characteristics of the target language, such as the diction.
We also find that the PPL of Moses is the largest, and its generated length reaches the length of the source text.
Moses seems to conduct word-to-word translation.
This practice is acceptable in text simplification, but totally offends the summarization requirement. 
Although not manually controlled, the lengths of the outputs in ABS and CoRe are both similar to the actual one, which demonstrates the learning ability of Seq2Seq models.
In addition, Table~\ref{tb:result} shows that compared to ABS, CoRe generates far fewer UNK tags and its copying ratio is closer to the actual one.
The former verifies the power of our two decoders, while the latter may be attributed to the supplement of the supervision of writing modes.


\subsection{Case Study}
\begin{table*}[ht]
	\centering
	\small
	\begin{tabularx}{0.9\linewidth}{p{1.1cm}|X}
		\hline
		Source    & another @entity34 military official who spoke on the condition of anonymity told @entity3 that the fall of @entity11 is not imminent \\ \hline
		Target & a \colorbox{yellow}{@entity34 military official} \colorbox{cyan}{tells} \colorbox{yellow}{@entity3} \colorbox{yellow}{the fall of @entity11 is not imminent}                                                   \\ 
		Moses     & \colorbox{yellow}{another @entity34 military official} \colorbox{yellow}{condition of anonymity} \colorbox{yellow}{told @entity3 the fall of @entity11 is not imminent}                                                                                                                                     \\ 
		ABS       &   the \colorbox{yellow}{@entity34 military official} \colorbox{yellow}{spoke on the condition of anonymity}                                                                                                                                 \\ 
		CoRe        & a \colorbox{yellow}{@entity34 military official} \colorbox{cyan}{said} \colorbox{yellow}{the fall of @entity11 is not imminent}                                                             \\ \hline
	\end{tabularx}
	\caption{Generation example in summarization. We use colors to distinguish the word source, i.e., \colorbox{yellow}{copying}, \colorbox{cyan}{alignment} or \colorbox{white}{common words}.}
	\label{tb:case}
\end{table*}
In addition to the automatic sentence quality measurements, we manually inspect what our model actually generates.
In text simplification, we observe that the paraphrase rules of Simple Wikipedia are relatively fixed.
For example, no matter how the article in Wikipedia illustrates, Simple Wikipedia usually adopts the following pattern to describe a commune:
\begin{quote}
	\#NAME is a commune . it is found in \#LOCATION .
\end{quote}
CoRe grasps many frequent paraphrase rules, and there are more than 130 cases where the generation results of CoRe exactly hit the actual target sentences.
Therefore, we focus more on the analysis of the summarization results next.
In summarization, although most target words come from the copying decoder, we find CoRe tends to pick keywords from different parts of the source document.
By contrast, the standard attentive Seq2Seq model often extracts a large part of continuous source words.
Meanwhile, the restricted generation decoder usually plays the role to ``connect'' these keywords, such as to change the tenses, or to supplement article words.
Its behavior resembles a human summarizer to a large extent.
Table~\ref{tb:case} gives some examples generated by different models.
We find that the sentence generated by CoRe is fluent and satisfies the need of summarization.
The only difference from the actual target is that CoRe does not assume ``told @entity3'' is important enough and simplifies it to ``said''.
It is the common way that human summarizes.
Notably, CoRe changes the starting word from ``another'' to ``a'', which is actually more preferred for an independent highlight.
Looking at other models, Moses almost repeats the content of the source text.
As a result, it is the longest one and fails to catch the main idea.
ABS indeed compresses the source text. It however focuses on the wrong place, i.e., the attributive clause.
Therefore, its output does not even form a complete sentence.

\section{Related Work}
The Seq2Seq model is a newly emerging approach.
It was initially proposed by \cite{kalchbrenner2013recurrent,sutskever2014sequence,cho2014properties} for machine translation.
Compared with the traditional statistical machine translation approaches (e.g., \cite{koehn2007moses}), Seq2Seq models require less human efforts.
Later, \cite{bahdanau2014neural} developed the attention mechanism which largely promoted the applications of the  Seq2Seq models.
In addition to machine translation, Seq2Seq models achieved the state-of-the-art performance in many other tasks such as response generation~\cite{shang2015neural}
Some researches (e.g., \cite{rush-chopra-weston:2015:EMNLP,hu2015lcsts}) have directly applied the general Seq2Seq model~\cite{bahdanau2014neural} to the paraphrase-oriented task.
However, the experiments of \cite{rush-chopra-weston:2015:EMNLP} demonstrated that the introduction of hand-crafted features significantly improved the performance of the original model.
Consequently, the general Seq2Seq model used for machine translation seemed not suitable for the paraphrase task which involves both copying and rewriting.

Limited work has explored the copying mechanism.
\cite{vinyals2015pointer} proposed a pointer mechanism to predict the output sequence directly from the input.
In addition to the different applications, their model cannot generate items outside of the set of input sequence.
Later, \cite{allamanis2016convolutional} developed a convolutional attention network to generate the function name of the source code.
Since there are many out-of-vocabulary (OOV) words in the source code, they used another attention model in the decoder to directly copy a code token.
In Seq2Seq generation, the most relevant work we find is COPYNET~\cite{gu2016incorporating} which has been explained in the introduction.

Some existing work has tried to modify the output dimension of the decoder to speed up the training process.
In training, \cite{cho2015using} restricted the decoder to generate the words from the actual target words together with a sampled word set.
\cite{nallapatiabstractive} supplemented the 1-nearest-neighbors of words in the source text, as measured by the similarity in the word embedding space.
Notice that, these models still decoded on the full vocabulary during test.
In comparison, our restricted generative decoder always produces the words in a small yet highly relevant vocabulary.

\section{Conclusion and Future Work}
In this paper, we develop a novel Seq2Seq model called CoRe to simulate the two core writing modes in paraphrase, i.e., copying and rewriting.
CoRe fuses a copying decoder and a restricted generative decoder.
To combine the two decoders and determine the final output, we train a predictor to predict the writing modes.
We conduct extensive experiments on two different paraphrase-oriented datasets.
The result shows that our model outperforms the state-of-the-art approaches in terms of both informativeness and language quality.
At present, our model focuses on producing a single sentence.
We plan to extend it to generate multi-sentence documents.

\section{ Acknowledgments}
The work described in this paper was supported by Research Grants Council of Hong Kong (PolyU 152094/14E), National Natural Science Foundation of China (61272291, 61672445) and The Hong Kong Polytechnic University (G-YBP6, 4-BCB5, B-Q46C).
The correspondence authors of this paper are Wenjie Li and Sujian Li.

\bibliographystyle{aaai}
\bibliography{aaai2016}

\begin{thebibliography}{}

\bibitem[\protect\citeauthoryear{Allamanis, Peng, and
  Sutton}{2016}]{allamanis2016convolutional}
Allamanis, M.; Peng, H.; and Sutton, C.
\newblock 2016.
\newblock A convolutional attention network for extreme summarization of source
  code.
\newblock {\em arXiv preprint arXiv:1602.03001}.

\bibitem[\protect\citeauthoryear{Bahdanau, Cho, and
  Bengio}{2014}]{bahdanau2014neural}
Bahdanau, D.; Cho, K.; and Bengio, Y.
\newblock 2014.
\newblock Neural machine translation by jointly learning to align and
  translate.
\newblock {\em arXiv preprint arXiv:1409.0473}.

\bibitem[\protect\citeauthoryear{Cho \bgroup et al\mbox.\egroup
  }{2014}]{cho2014properties}
Cho, K.; Van~Merri{\"e}nboer, B.; Bahdanau, D.; and Bengio, Y.
\newblock 2014.
\newblock On the properties of neural machine translation: Encoder-decoder
  approaches.
\newblock {\em arXiv preprint arXiv:1409.1259}.

\bibitem[\protect\citeauthoryear{Cho, Memisevic, and
  Bengio}{2015}]{cho2015using}
Cho, S. J.~K.; Memisevic, R.; and Bengio, Y.
\newblock 2015.
\newblock On using very large target vocabulary for neural machine translation.

\bibitem[\protect\citeauthoryear{Dyer, Chahuneau, and
  Smith}{2013}]{dyer2013simple}
Dyer, C.; Chahuneau, V.; and Smith, N.~A.
\newblock 2013.
\newblock A simple, fast, and effective reparameterization of ibm model 2.
\newblock Association for Computational Linguistics.

\bibitem[\protect\citeauthoryear{Gu \bgroup et al\mbox.\egroup
  }{2016}]{gu2016incorporating}
Gu, J.; Lu, Z.; Li, H.; and Li, V.~O.
\newblock 2016.
\newblock Incorporating copying mechanism in sequence-to-sequence learning.
\newblock {\em arXiv preprint arXiv:1603.06393}.

\bibitem[\protect\citeauthoryear{Hermann \bgroup et al\mbox.\egroup
  }{2015}]{hermann2015teaching}
Hermann, K.~M.; Kocisky, T.; Grefenstette, E.; Espeholt, L.; Kay, W.; Suleyman,
  M.; and Blunsom, P.
\newblock 2015.
\newblock Teaching machines to read and comprehend.
\newblock In {\em Advances in Neural Information Processing Systems},
  1693--1701.

\bibitem[\protect\citeauthoryear{Hu, Chen, and Zhu}{2015}]{hu2015lcsts}
Hu, B.; Chen, Q.; and Zhu, F.
\newblock 2015.
\newblock Lcsts: A large scale chinese short text summarization dataset.
\newblock {\em arXiv preprint arXiv:1506.05865}.

\bibitem[\protect\citeauthoryear{Kalchbrenner and
  Blunsom}{2013}]{kalchbrenner2013recurrent}
Kalchbrenner, N., and Blunsom, P.
\newblock 2013.
\newblock Recurrent continuous translation models.
\newblock In {\em EMNLP}, volume~3,  413.

\bibitem[\protect\citeauthoryear{Kauchak}{2013}]{kauchak2013improving}
Kauchak, D.
\newblock 2013.
\newblock Improving text simplification language modeling using unsimplified
  text data.
\newblock In {\em ACL (1)},  1537--1546.

\bibitem[\protect\citeauthoryear{Koehn \bgroup et al\mbox.\egroup
  }{2007}]{koehn2007moses}
Koehn, P.; Hoang, H.; Birch, A.; Callison-Burch, C.; Federico, M.; Bertoldi,
  N.; Cowan, B.; Shen, W.; Moran, C.; Zens, R.; et~al.
\newblock 2007.
\newblock Moses: Open source toolkit for statistical machine translation.
\newblock In {\em Proceedings of the 45th annual meeting of the ACL on
  interactive poster and demonstration sessions},  177--180.
\newblock Association for Computational Linguistics.

\bibitem[\protect\citeauthoryear{Lin}{2004}]{lin2004rouge}
Lin, C.-Y.
\newblock 2004.
\newblock Rouge: A package for automatic evaluation of summaries.
\newblock In {\em Proceedings of the ACL Workshop},  74--81.

\bibitem[\protect\citeauthoryear{Nallapati \bgroup et al\mbox.\egroup
  }{}]{nallapatiabstractive}
Nallapati, R.; Zhou, B.; glar Gul{\c{c}}ehre, {\c{C}}.; and Xiang, B.
\newblock Abstractive text summarization using sequence-to-sequence rnns and
  beyond.

\bibitem[\protect\citeauthoryear{Rush, Chopra, and
  Weston}{2015}]{rush-chopra-weston:2015:EMNLP}
Rush, A.~M.; Chopra, S.; and Weston, J.
\newblock 2015.
\newblock A neural attention model for abstractive sentence summarization.
\newblock In {\em Proceedings of EMNLP},  379--389.

\bibitem[\protect\citeauthoryear{Shang, Lu, and Li}{2015}]{shang2015neural}
Shang, L.; Lu, Z.; and Li, H.
\newblock 2015.
\newblock Neural responding machine for short-text conversation.
\newblock {\em arXiv preprint arXiv:1503.02364}.

\bibitem[\protect\citeauthoryear{Sutskever, Vinyals, and
  Le}{2014}]{sutskever2014sequence}
Sutskever, I.; Vinyals, O.; and Le, Q.~V.
\newblock 2014.
\newblock Sequence to sequence learning with neural networks.
\newblock In {\em Advances in neural information processing systems},
  3104--3112.

\bibitem[\protect\citeauthoryear{Tieleman and
  Hinton}{2012}]{tieleman2012lecture}
Tieleman, T., and Hinton, G.
\newblock 2012.
\newblock Lecture 6.5-rmsprop: Divide the gradient by a running average of its
  recent magnitude.
\newblock {\em COURSERA: Neural Networks for Machine Learning} 4(2).

\bibitem[\protect\citeauthoryear{Vinyals, Fortunato, and
  Jaitly}{2015}]{vinyals2015pointer}
Vinyals, O.; Fortunato, M.; and Jaitly, N.
\newblock 2015.
\newblock Pointer networks.
\newblock In {\em Advances in Neural Information Processing Systems},
  2692--2700.

\end{thebibliography}

\end{document}